\documentclass[runningheads]{llncs}
\usepackage{graphicx}
\usepackage{amssymb}
\usepackage{amsmath}
\usepackage{xcolor}
\usepackage{booktabs}
\usepackage{tabularx}
\usepackage{graphicx}
\usepackage{hyperref}
\usepackage{subfig}

\begin{document}
\title{Evaluating the Fairness of Neural Collapse in Medical Image Classification}

%
\titlerunning{Fairness of Neural Collapse}
%
\author{Kaouther Mouheb\thanks{Corresponding author: k.mouheb@erasmusmc.nl}\inst{1} \and
Marawan Elbatel \inst{2} \and
Stefan Klein \inst{1} \and 
Esther E. Bron \inst{1}}

\authorrunning{K. Mouheb et al.}
%
\institute{Biomedical Imaging Group Rotterdam, Department of Radiology \& Nuclear Medicine, Erasmus MC, the Netherlands
\and
The Hong Kong University of Science and Technology, Hong Kong, China
\\
}
\maketitle              
\begin{abstract}

Deep learning has achieved impressive performance across various medical imaging tasks. However, its inherent bias against specific groups hinders its clinical applicability in equitable healthcare systems. A recently discovered phenomenon, Neural Collapse (NC), has shown potential in improving the generalization of state-of-the-art deep learning models. Nonetheless, its implications on bias in medical imaging remain unexplored. Our study investigates deep learning fairness through the lens of NC. We analyze the training dynamics of models as they approach NC when training using biased datasets, and examine the subsequent impact on test performance, specifically focusing on label bias. We find that biased training initially results in different NC configurations across subgroups, before converging to a final NC solution by memorizing all data samples. Through extensive experiments on three medical imaging datasets—PAPILA, HAM10000, and CheXpert—we find that in biased settings, NC can lead to a significant drop in F1 score across all subgroups. Our code is available at \href{https://gitlab.com/radiology/neuro/neural-collapse-fairness}{https://gitlab.com/radiology/neuro/neural-collapse-fairness}.
\end{abstract}
\section{Introduction}

Recent progress in medical image analysis has been greatly shaped by new deep learning (DL) models. Enhanced hardware capabilities enabled training over-parameterized models, allowing them to achieve comparable performance to practicing radiologists in some cases \cite{rajpurkar2018deep}. Although effective, integrating these techniques into clinical practice is impeded by social and ethical concerns \cite{kelly2019key}. DL-based diagnostic tools often face fairness issues as they display biases toward demographic groups based on race, age, sex, and other factors, undermining the goal of equitable healthcare systems \cite{mehta2024evaluating,chen2023algorithmic}. 
Neural Collapse (NC) is a notable development in DL research.  Papyan et al. \cite{papyan2020prevalence} show its potential to enhance model robustness, interpretability, and generalization. NC-inspired techniques emerged in various DL domains, such as imbalanced learning and federated learning \cite{Li_2023_ICCV,xie2023neural,zhu2023bridging}. However, recent studies highlight NC's limited test generalization, calling for further investigation \cite{hui2022limitations}. In this regard, the impact of NC on model fairness and performance under biased training scenarios remains unexplored.

Multiple algorithmic methods have emerged as solutions to DL bias issues \cite{lu2022importance,zikang2023fairadabn,yuan2023embarass}. However, the MEDFAIR Benchmarking framework \cite{zong2022medfair} revealed their limited efficiency compared to traditional learning approaches such as Empirical Risk Minimization (ERM).
Moreover, there is currently no standard metric to assess fairness, rendering the evaluation of these techniques even more challenging \cite{mbakwe2023fairness}. These issues prompt the need for a deeper understanding of bias emergence in DL models, in order to design efficient bias mitigation methods and better fairness metrics \cite{jones2023role}. In this context, Jones et al. \cite{jones2023role} revealed that models trained with biased datasets can encode sensitive information about the subgroups in their extracted features, leading to inter-group performance disparities.
The NC phenomenon discovered by Papyan et al. \cite{papyan2020prevalence} is a compelling empirical state observed in over-parameterized models trained beyond zero training error. NC occurs when the intra-class variability of the extracted features approaches zero, while their class means form a symmetric geometric structure called a simplex equiangular tight frame (ETF).
Under this definition, it is asserted that as models approach NC, features extracted from samples of the same label converge to the same representation, irrespective of subgroup differences. Consequently, a pertinent question arises: \textit{does this convergence facilitate the attenuation of sensitive information embedded in the features? and what are its implications on test performance across distinct population subgroups?}

In light of the aforementioned work, this study addresses this inquiry by examining the fairness of medical image classification models through NC. It aims to bridge the existing gap in understanding the impact of NC on model performance across subgroups under biased training, focusing on standard training with ERM. The contributions of this work can be summarized as follows: \textbf{(i)} We analyze NC properties under biased training, focusing on label bias \textbf{(ii)} we show that models approaching NC appear to encode less sensitive subgroup information in the extracted features \textbf{(iii)} we empirically demonstrate degraded performance across all subgroups upon convergence to NC under biased training.

\section{Preliminaries}

Following the work of Jones et al. \cite{jones2023role}, we focus on the bias stemming from under-diagnosis within a binary classification framework. The task is to build a model that classifies samples into either ``positive" denoting the presence of a disease, or ``negative" denoting its absence.
Consider a dataset $D = \left \{ \left ( x_{i}, a_{i}, y_{i} \right ) \right \}_{i=1}^{n}$ of $n$ samples. Each sample $x\in \mathbb{R}^{d}$ is associated with a binary label $y \in Y : \left \{ y^+, y^- \right \}$ and belongs to a subgroup $a \in A$. 
The training set is biased against a group $a^*$ when its distribution inaccurately represents that group's characteristics, leading to skewed model predictions. 
In the case of under-diagnosis, individuals from the positive class in group $a^*$ are mistakenly labeled as negative.

Neural collapse is defined as a state where the outputs of the last feature extraction layer converge towards their intra-class means. Simultaneously, these class means and the weights of the linear classifier converge towards the vertices of a simplex ETF \cite{papyan2020prevalence}. In practice, models do not exactly attain NC, but they approach it as the training progresses \cite{sukenik2023deep}. The NC configuration of a model is defined by the class means of its features and its linear classifier's weights. The optimal NC configuration is characterized by four properties: \\
\textbf{NC1: Variability collapse:} Intra-class variability of the last layer features approaches zero as the features converge to the corresponding class means: 
\begin{equation}
   S = \frac{1}{n} \sum_{i=1}^{n} \left \| h_{i,k} - \mu_k \right \|_2 \rightarrow 0
\label{eq_nc1}
\end{equation}
where $h_{i,k}$ is the feature representation of the $i^{th}$ sample and $k$ is its class label, $\mu_k = \frac{1}{n_k}\sum_{i=1}^{n_k}h_{i,k} $ is the mean of the $k^{th}$ class with a number of samples $n_k$.\\
\textbf{NC2: Convergence to a simplex ETF:} The vectors defined by the class means $\mu_k$ converge to the vertices of a geometric structure where each pair of vectors have equal lengths and are positioned at equal angles from each other: 

\[ \begin{aligned}
\left | \left \| \mu_k - \mu_G \right \|_2 - \left \| \mu_{k'} - \mu_G \right \|_2 \right |\rightarrow 0 ~~ \forall ~ k, k' 
\\  
\left \langle \widetilde{\mu}_k, ~ \widetilde{\mu}_{k'} \right \rangle \rightarrow \frac{K}{K-1}\delta_{k, k'} -\frac{1}{K-1} ~~ \forall ~ k,k' 
\end{aligned}\]  
$\mu_G = \frac{1}{K} \sum_{k=1}^{K} \mu_k$ is the global mean and $\widetilde{\mu}_k = \left ( \mu_k - \mu_G \right ) / \left\| \mu_k - \mu_G \right\|_2$, $K$ is the number of classes and $\delta_{k, k'}$ is the Kronecker delta operator. \\
\textbf{NC3: Convergence to self-duality:} The class means $\mu_k$ and the weights of the linear classifier $\mathrm{w}_k$ converge to the same simplex ETF (up to re-scaling), $\widetilde{\mu}_k = \frac{\mathrm{w}_k}{\left \| \mathrm{w}_k \right \|_F}$. $F$ refers to the Frobenius norm. \\
\textbf{NC4: Simplification to a nearest class center predictor:} The linear classifier of the model assigns each sample to the class with the closest mean, 
$\mathrm{argmax}_k \left \langle h,w_k \right \rangle \rightarrow \mathrm{argmin}_k \left \| h-\mu_k \right \|$.

\section{Neural Collapse Under Biased Training}
To investigate the theoretical potential of NC in training fair deep classification models, we examine the impact of biased training on the feature encoding process in the context of NC. While NC2-4 relate to class means and classifier weights shared across groups, NC1 involves individual samples. Thus, we focus on NC1 to analyze how samples from each group converge towards their class mean. Taking the subgroups into consideration, equation \ref{eq_nc1} can be formulated as follows:
\begin{equation}
    S = \sum_{a\in A} S_a = \sum_{a\in A} \frac{1}{n_a} \sum_{i=0}^{n_a} \left \| h_{i,k} - \mu_k \right \|_2 \rightarrow 0
    \label{eq_nc1_a}
\end{equation}
where $n_a$ is the number of samples belonging to group $a$. Equation \ref{eq_nc1_a} implies: 
\begin{equation}
    S_a =\frac{1}{n_a} \sum_{i=0}^{n_a} \left \| h_{i,k} - \mu_k \right \|_2 \rightarrow 0 ~~~ \forall a \in A
    \label{eq_nc1_a_tpt}
\end{equation}

In an unbiased training scenario, supervised training with ERM drives all samples towards the vertices of the simplex ETF defined by NC2 as the model approaches NC. The model in this case learns the same mapping for all groups $P(h_{i,k}|x_i, a) ~~ \forall a \in A$. Previous research found that in the presence of label noise, models initially focus on fitting clean samples before memorizing the noisy ones \cite{liu2020early}. This  allows us to analyze the training process in two phases (Figure  \ref{fig:visual_example}):
\begin{figure}[t]
    \centering
    \includegraphics[width=\textwidth]{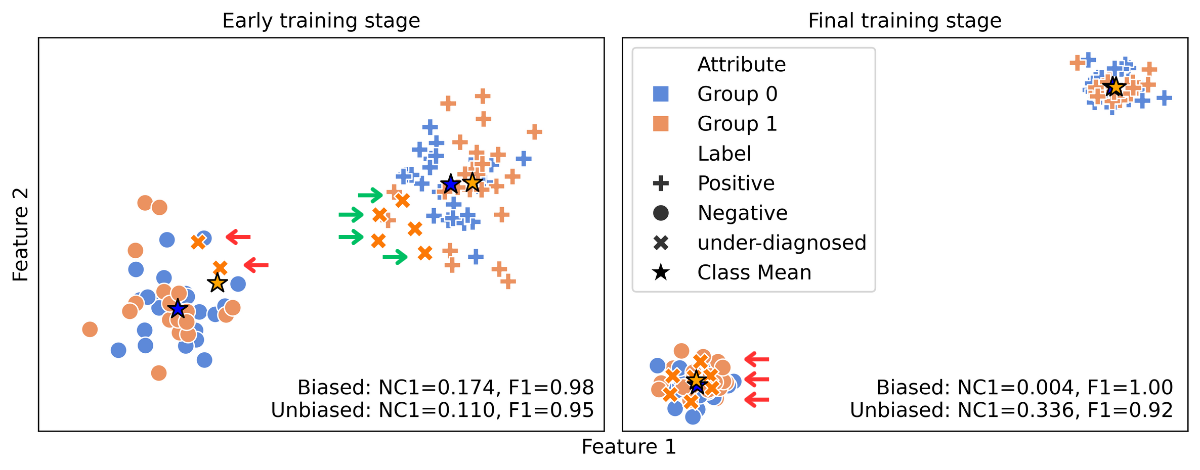}
    \caption{A 2D example of variability collapse under label noise. The crosses (x) are positive samples (+) from Group 1 (orange) that are mistakenly classified as negative samples (-). In early  training stages the majority of them are close to the positive class mean (right arrows) leading to poor train NC but a high performance on unbiased data. The final phase of training drives all noisy samples closer to the negative class mean (left arrows) leading to an optimal train collapse but a drop in test performance (Colored figure available online).}
    
    \label{fig:visual_example}
\end{figure}
\\
\textbf{Early Training Stage:} According to Nguyen et al. \cite{nguyen2022memorization}, the model first learns distinct NC configurations for the clean and noisy samples. In the biased setting, the label noise affects a specific population group $a^*$. This implies that the model learns a distinct NC configuration for the  under-diagnosed group $a^*$: 
\begin{equation}
\left [ \mu_{0,a^*}, \mu_{1,a^*} \right ] \neq \left[ \mu_{0,a'}, \mu_{1,a'} \right ] ~~ \forall a' \neq a^*
\end{equation} 
Besides, the model learns feature extraction primarily from clean labels. Since samples coming originally from the same class tend to exhibit similar input characteristics, the under-diagnosed samples are mapped closer to the positive class mean $\mu_1$ during this phase. Thus, the model diverges from its optimal NC configuration leading to slower NC convergence. Nonetheless, performance on an unbiased test set improves since the features are learned from clean data.\\
\textbf{Final Training Stage:} According to equation \ref{eq_nc1_a_tpt}, as models reach the final stage, samples with the same label $k$ converge to identical feature representations $\mu_k$, yielding $S_a \approx S_{a'} \approx 0 ~~ \forall a,a' \in A$. Hence, theoretically, all groups attain identical NC configurations, rendering samples from different groups indistinguishable at the feature level. To attain NC, the model overfits the data, driving the mislabeled samples closer to to the negative class mean $\mu_0$. Thus, inputs with similar characteristics are embedded to maximally separated features (vertices of a simplex ETF), causing inconsistency in the model's feature encoding process. Consequently, a degradation is expected in the test performance of all subgroups.

\section{Experimental Results}
\subsection{Experimental Setting}
We conduct experiments on three public medical imaging datasets, namely PAPILA, HAM10000, and CheXpert, spanning three modalities: fundus, dermatoscopic, and X-ray imaging \cite{kovalyk2022papila,tschandl2018ham10000,irvin2019chexpert}. All labels are converted to binary (0 for healthy, 1 for unhealthy samples).
We explore two demographic attributes in each dataset, with a total of six dataset-attribute combinations. Table \ref{tab:datasets} gives an overview of the datasets. 
We follow the framework of Jones et al. \cite{jones2023role} where for each combination, a  model is trained on a clean and a biased set. In the biased set, randomly selected 25\% of positive samples in Group 1 are mislabeled as negative in the train and validation sets. Both models are tested on the same unbiased test set. We compare models trained for 200 epochs to models saved during the initial training phase via early stopping. Experiments are repeated for 10 random seeds. Implementation details are provided in Appendix A.2. 
\begin{table}[t]
\caption{Demographic distributions of the datasets. G0 refers to Group 0 and G1 refers to Group 1. The numbers in parentheses represent the number/percentage of the positive samples. Splits are shown in train, validation, test order.}
\label{tab1}
\centering
\begin{tabularx}{\textwidth}{X X X c}
\toprule
 & \textbf{PAPILA} & \textbf{HAM10000} & \textbf{CheXpert}\\
\midrule
\textbf{Samples} & 420 (87) & 9958 (1438) & 127118 (116202)  \\
\textbf{Splits (\%)} & 70-10-20 & 80-10-10 & 60-10-30 \\
\midrule
\textbf{G0: Male} & 34.8\% (24.0\%) & 54.2\% (16.8\%) & 58.8\% (91.6\%) \\
\textbf{G1: Female} & 65.2\% (19.0\%) & 45.8\% (11.6\%) & 41.2\% (91.2\%)\\
\midrule
\textbf{G0: Age $<$ 60} &  40.5\% (6.47\%) &  71.9\% (9.55\%) & -\\
\textbf{G1: Age $\geq$ 60} & 59.5\% (30.4\%) & 28.1\% (26.9\%) & - \\
\midrule
\textbf{G0: White} & - & - & 77.9\% (91.7\%) \\
\textbf{G0: Non-White} & - & - & 22.1\% (90.5\%) \\
\bottomrule
\end{tabularx}
\label{tab:datasets}
\end{table}
\subsection{Neural Collapse Convergence Under Label Bias}

We monitor the NC properties (NC1-4) of each model during training; see Appendix A.1 for detailed metrics \cite{kothapalli2022neural,sukenik2023deep}. Figure \ref{fig:nc1_convergence} illustrates the NC1 metric plots, while the plots for NC2-4 can be found in Appendix A.3.

The results align with our analysis, as models trained under label bias exhibit elevated NC1 values during the initial phases, suggesting a tendency to prioritize clean samples while pushing the under-diagnosed samples farther from the negative class mean. As training progresses, both models approach zero train NC1, indicating that all samples, including mislabeled ones, are memorized by the model. In CheXpert, the discrepancy is more pronounced, likely because the positive class, in which the bias is injected, constitutes 91\% of the dataset, leading to a higher proportion of mislabeled samples compared to the other sets.

\begin{figure}[t]
    \centering
    \includegraphics[width=\textwidth]{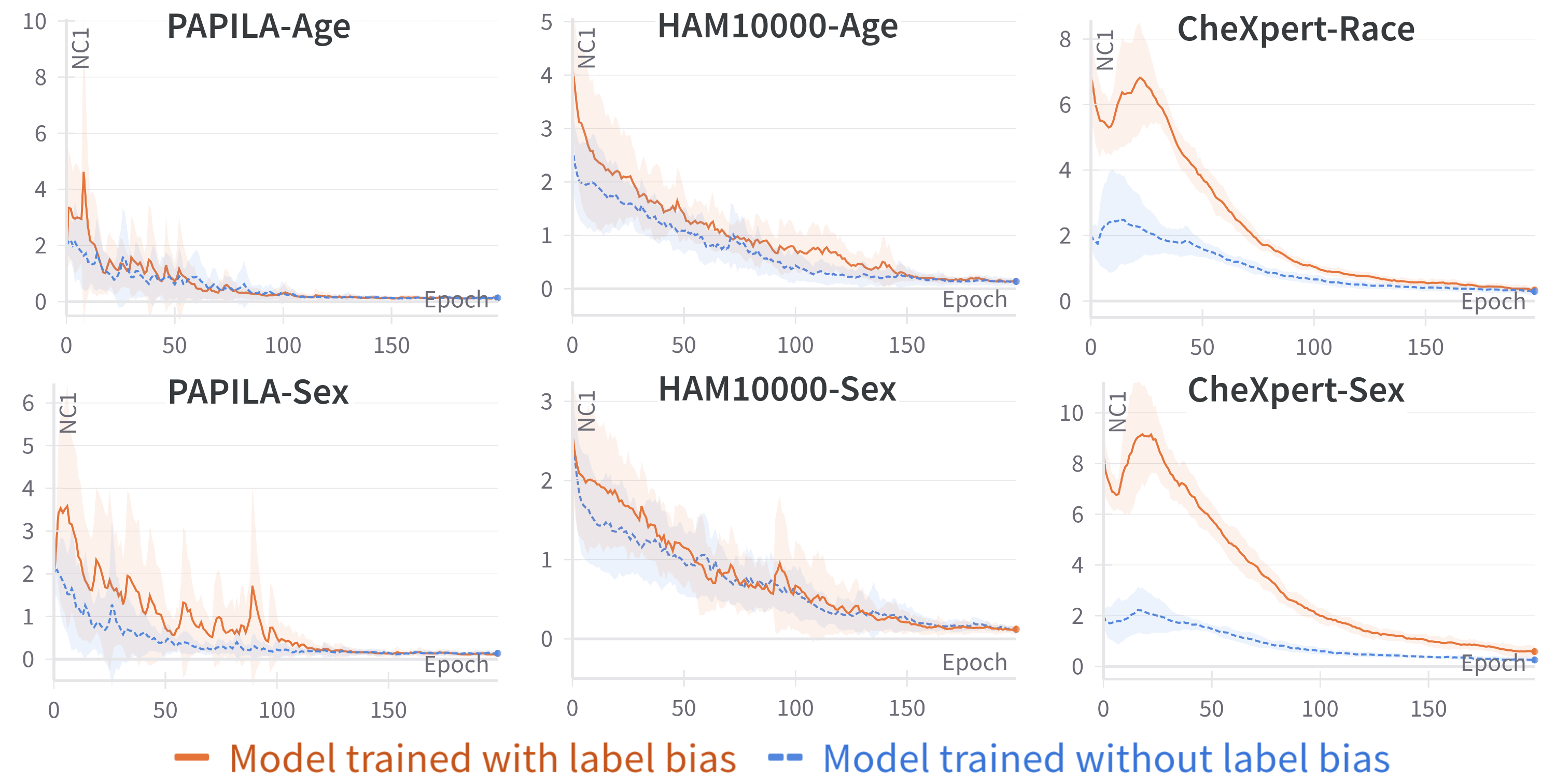}
    \caption{NC1 metric per epoch for each dataset-attribute combination. Biased training (solid orange line) exhibits higher initial NC1 values and slower convergence to NC compared to unbiased training (dashed blue line). Shaded areas represent the standard deviation across 10 random seeds.}\label{fig:nc1_convergence}
\end{figure}

\subsection{Feature-Level Group Separability of NC Solutions}
\begin{figure}[t]
    \centering
    \includegraphics[width=\textwidth]{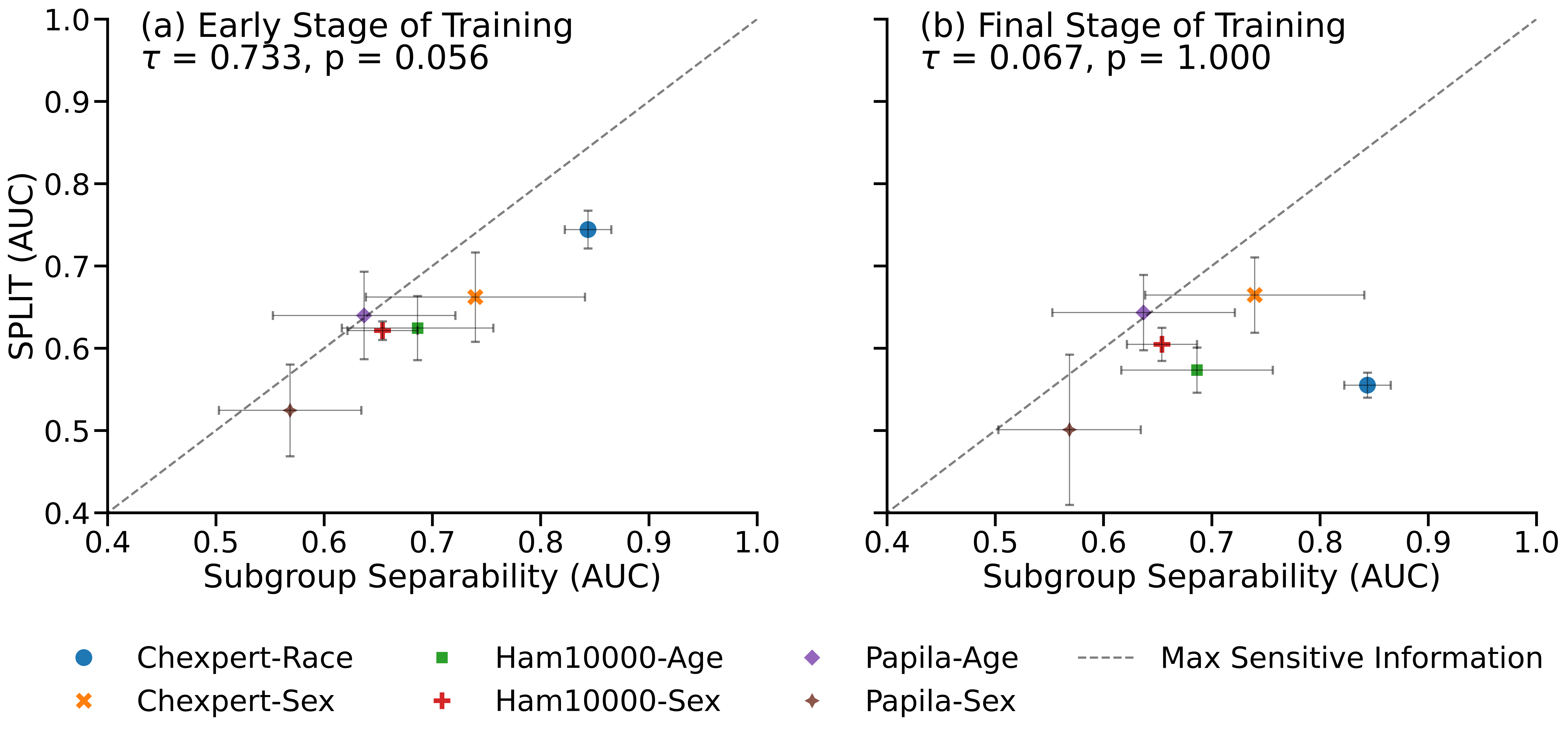}
    \caption{AUC of the SPLIT test for sensitive information encoded in extracted features against subgroup separability of the raw data. While data points in early stage training (a) are on the y=x axis, this is not found in the final stage of training (b), indicating that models closer to NC remove group information. Error bars represent the standard deviation across 10 random seeds.} 
    \label{fig:split_results}
\end{figure}

We compare the amount of group information encoded in the features of the biased model to that present in the images using the Supervised Prediction Layer Information Test (SPLIT) \cite{glocker2023algorithmic,groh2022towards}: A linear classifier is trained to predict the attributes from the features extracted by the disease classification models. A model is trained to predict the attributes from the raw images to measure the group information in the data. We plot the AUC of the SPLIT test against the AUC of this model for the early and final stage models. Kendall’s $\tau$ statistic is used to assess the monotonic association between these AUCs (Figure \ref{fig:split_results}).

The Kendall’s $\tau$ statistic applied to early-stage features suggests that the models encode nearly as much group information in their features as the raw images. In the later stage, models approaching NC appear to encode reduced subgroup information (see CheXpert-Race), where the lower AUC indicates that samples belonging to different groups become indistinguishable at the feature level. However, high scores are still observed in some experiments such as CheXpert-Sex. This shows that in practice, the model's NC convergence highly depends on the training data, where at 200 epochs, some models still map samples from different groups to distinct NC configurations. 

\subsection{Test-Time Generalization of NC Solutions}
\begin{figure}[t]
\includegraphics[width=\textwidth]{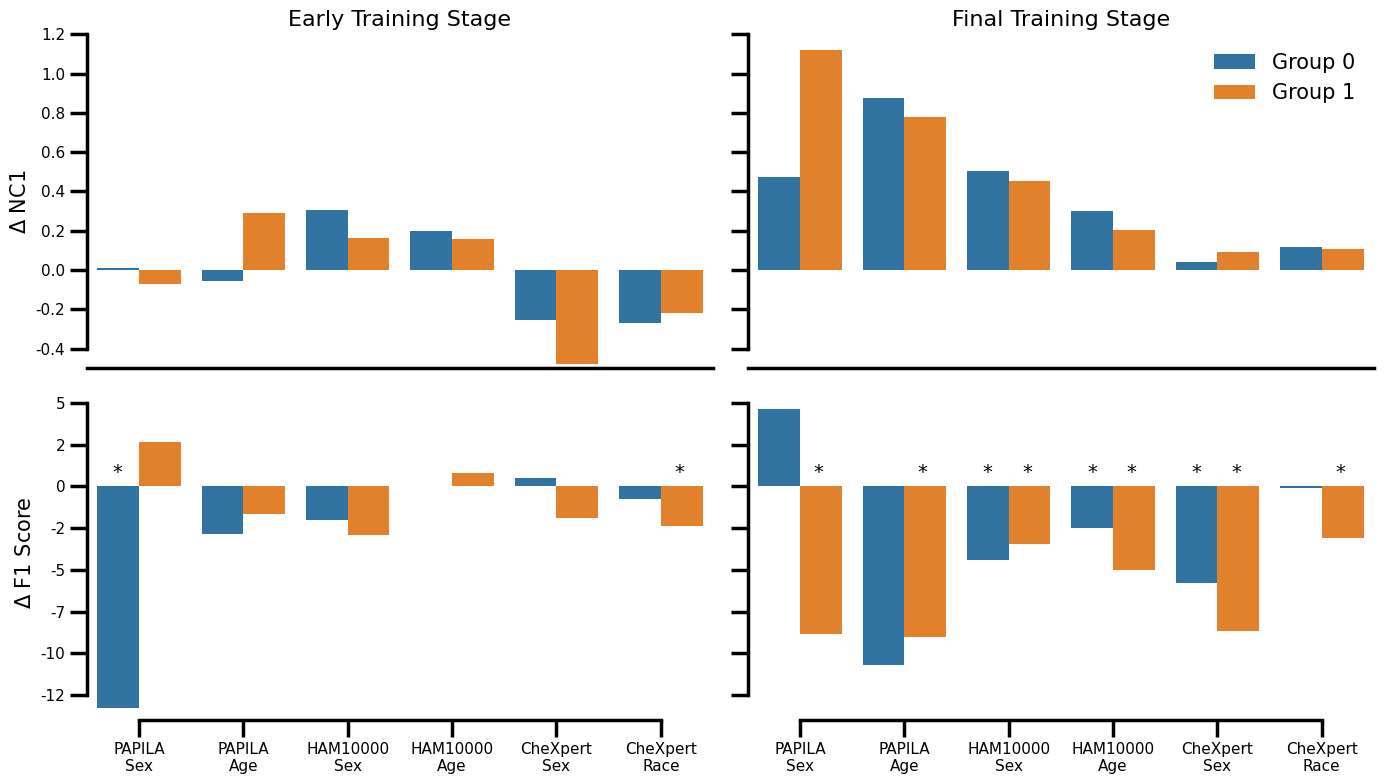}
\caption{Test-time differences in NC1 and F1 scores between biased and unbiased models. A positive value of $\Delta$NC1 means the biased model exhibits a higher NC1 reflecting a worse test NC. A negative value of $\Delta$F1 score indicates that the biased model achieves worse F1 score compared to the model trained with clean data. The * denotes statistically significant F1 score difference.} \label{fig:test_generalization}
\end{figure}

To assess the generalization of NC solutions on unbiased test data, we compare the results of models trained on biased data to those trained on clean data. We report the test NC1 and F1 score for each group. We examine the results of the models saved during initial training stages and those trained for 200 epochs. We test the statistical significance of F1 score differences using a Mann-Whitney U test with a $p_{critical} = 0.05$ (Figure \ref{fig:test_generalization}).

The results show that during early stages, models exhibit no significant difference in F1 scores across most dataset-attribute combinations, indicating reliance on the clean data for feature extraction. Exceptions occur in minority groups, namely males in PAPILA (34.8\%) and non-whites in CheXpert (22.1\%). This highlights the model's tendency to learn distinct NC configurations for different groups, where the small size of effective training data for these groups led to poor performance.
In the final stage, while all models approach zero train NC1  (Figure \ref{fig:nc1_convergence}), biased models show increased test NC1 for all groups compared to the clean models. This is more pronounced in smaller datasets, as they are easier to overfit, making the feature encoding process more inconsistent. Consequently, a significant F1 score gap is observed between the biased and clean models. The under-diagnosed group (Group 1) consistently suffers performance degradation, while Group 0 is negatively affected in four out of six experiments. Interestingly, although the model seems to use less subgroup information in the CheXpert-Race experiment (Figure \ref{fig:split_results}), a difference in test performance is seen between the subgroups in the final stage. While it is important to treat the statistical significance of the F1 score difference with caution due to the small sample size (10 experiments), a possible explanation is that in practice, since NC is not exactly attained, the class means are biased towards the white population due its larger number of samples, resulting in performance disparities.

\section{Discussion}
This study evaluates the effects of biased training on the feature encoding of medical image classification models through the phenomenon of Neural Collapse (NC), and its implications on model generalization, with a specific focus on label bias.  Our experiments highlight the two-phase training process under biased conditions. Initially, models learn to encode features from clean data before incorporating noisy samples as the model converges to NC. This impedes the feature encoding process since samples originally coming from the same class are mapped to maximally separable representations. Additionally, our findings suggest that convergence to NC can reduce group information in the extracted features, however, this is usually not attained in practice. Finally, we show that approaching train NC does not guarantee test collapse in biased settings. The inconsistency in the feature encoding during the final stages leads to poorer test NC and consequently degraded test performance in all subgroups. 

In essence, this paper offers initial insights into the complex interplay between biased training and NC. We present an NC-based analysis of the mechanics behind the emergence of bias in deep classification models and the consequent degradation in performance that occurs upon convergence to the NC solution. We hence emphasize the importance of taking fairness issues into consideration when developing NC-inspired solutions, especially in medical imaging, where dataset biases are prevalent.

We limited the scope of this study to binary classification with two population subgroups and a bias level of 25\%. Future work will extend this to include multiple population subgroups and different bias levels. Additionally, we plan to investigate multi-class classification tasks for fair differential disease diagnosis. We will also explore different bias sources and evaluate fairness in 3D modalities such as MRI and CT scans. Finally, future work will examine the effects of advanced bias and noise mitigation techniques on NC convergence, compared to the standard training with ERM. 
Through these efforts, we aim to refine our understanding and improve the fairness and reliability of deep learning in medical image analysis.

\begin{credits}

\subsubsection{\ackname} This project is supported by a 2022 Erasmus MC Fellowship. Esther E. Bron is recipient of TAP-dementia, a ZonMw funded project (\#10510032120003) in the context of the Dutch National Dementia Strategy. Esther E. Bron and Stefan Klein are recipients of EUCAIM, Cancer Image Europe, co-funded by the European Union under Grant Agreement 101100633. Marawan Elbatel is supported by the Hong Kong PhD Fellowship Scheme (HKPFS) from the Hong Kong Research Grants Council.

\subsubsection{\discintname}
The authors have no competing interests to declare that are
relevant to the content of this article.
\end{credits}

\bibliographystyle{splncs04}
\bibliography{main}

\end{document}